\PassOptionsToPackage{table}{xcolor}
\documentclass[11pt]{article}

\usepackage[preprint]{acl}

\usepackage{times}
\usepackage{latexsym}
\usepackage[T1]{fontenc}
\usepackage[utf8]{inputenc}
\usepackage{microtype}
\usepackage{inconsolata}
\usepackage{graphicx}
\usepackage{booktabs}
\usepackage{amsmath}
\usepackage{amssymb}
\usepackage{listings}
\usepackage{arydshln}
\usepackage{algorithm}
\usepackage{algpseudocode}
\usepackage{multirow}
\usepackage{soul} 
\newcommand{\hlerr}[1]{\textcolor{red}{\bfseries #1}}

\definecolor{codebg}{RGB}{255, 250, 230}
\definecolor{codeborder}{RGB}{180, 160, 80}
\definecolor{rulebg}{RGB}{230, 242, 255}
\definecolor{ruleborder}{RGB}{80, 130, 180}
\definecolor{exbg}{RGB}{255, 240, 245}
\definecolor{exborder}{RGB}{200, 120, 160}
\definecolor{diagbg}{RGB}{250, 245, 255}    
\definecolor{diagborder}{RGB}{140, 120, 180} 
\definecolor{taskbg}{RGB}{245, 250, 240}
\definecolor{taskborder}{RGB}{120, 160, 120}

\lstdefinestyle{taskbox}{
    basicstyle=\ttfamily\scriptsize,
    breaklines=true,
    columns=fullflexible,
    keepspaces=true,
    backgroundcolor=\color{taskbg},
    frame=single,
    framerule=0.8pt,
    rulecolor=\color{taskborder},
    framesep=6pt,
    xleftmargin=0.5em,
    xrightmargin=0.5em,
    aboveskip=1em,
    belowskip=1em,
}

\lstdefinestyle{ebnf}{
  basicstyle=\ttfamily\footnotesize,
  breaklines=true,
  breakatwhitespace=true,
  columns=fullflexible,
  keepspaces=true,
  showstringspaces=false,
  backgroundcolor=\color{codebg},
  frame=tblr,
  framerule=0.8pt,
  rulecolor=\color{codeborder},
  framesep=6pt,
  xleftmargin=0.5em,
  xrightmargin=0.5em,
  aboveskip=1em,
  belowskip=1em
}

\lstdefinestyle{rules}{
  basicstyle=\ttfamily\footnotesize,
  breaklines=false,
  columns=fullflexible,
  keepspaces=true,
  showstringspaces=false,
  backgroundcolor=\color{rulebg},
  frame=tblr,
  framerule=0.8pt,
  rulecolor=\color{ruleborder},
  framesep=6pt,
  xleftmargin=0.5em,
  xrightmargin=0.5em,
  aboveskip=1em,
  belowskip=1em
}

\lstdefinestyle{example}{
  basicstyle=\ttfamily\footnotesize,
  breaklines=true,
  breakatwhitespace=true,
  columns=fullflexible,
  keepspaces=true,
  showstringspaces=false,
  backgroundcolor=\color{exbg},
  frame=tblr,
  framerule=0.8pt,
  rulecolor=\color{exborder},
  framesep=6pt,
  xleftmargin=0.5em,
  xrightmargin=0.5em,
  aboveskip=1em,
  belowskip=1em,
}

\lstdefinestyle{errbox}{
    basicstyle=\ttfamily\footnotesize,
    breaklines=true,
    columns=fullflexible,
    keepspaces=true,
    backgroundcolor=\color{diagbg},
    frame=tblr,
    framerule=0.8pt,
    rulecolor=\color{diagborder},
    framesep=6pt,
    xleftmargin=0.5em,
    xrightmargin=0.5em,
    aboveskip=1em,
    belowskip=1em,
}

\title{Diagnosing CFG Interpretation in LLMs}

\author{
  Hanqi Li$_{1,4,5}$ \and
  Lu Chen$_{1,3,4,5}$ \and
  Kai Yu$_{1,2,4,5}$\thanks{Corresponding author.} \\[4pt]
  $^1$X-LANCE Lab, School of Computer Science, Shanghai Jiao Tong University, Shanghai, China \\
  $^2$AISpeech Co., Ltd., Suzhou, China \\
  $^3$Shanghai Innovation Institution, Shanghai, China \\
  $^4$Jiangsu Key Lab of Language Computing, Suzhou, China \\
  $^5$Suzhou Laboratory, Suzhou, China \\[2pt]
  \texttt{daqige@sjtu.edu.cn, chenlusz@sjtu.edu.cn, kai.yu@sjtu.edu.cn}
}

\begin{document}
\maketitle

\begin{abstract}
As LLMs are increasingly integrated into agentic systems, they must adhere to dynamically defined, machine-interpretable interfaces. We evaluate LLMs as in-context interpreters: \textit{given a novel context-free grammar, can LLMs generate syntactically valid, behaviorally functional, and semantically faithful outputs?} We introduce \textsc{RoboGrid}, a framework that disentangles \textbf{syntax}, \textbf{behavior}, and \textbf{semantics} through controlled stress-tests of recursion depth, expression complexity, and surface styles. Our experiments reveal a consistent hierarchical degradation: LLMs often maintain surface syntax but fail to preserve structural semantics. Despite the partial mitigation provided by CoT reasoning, performance collapses under structural density, specifically deep recursion and high branching, with semantic alignment vanishing at extreme depths. Furthermore, "\texttt{Alien}" lexicons reveal that LLMs rely on semantic bootstrapping from keywords rather than pure symbolic induction. These findings pinpoint critical gaps in hierarchical state-tracking required for reliable, grammar-agnostic agents.
\end{abstract}
\vspace{-10pt}
\begin{figure*}
    \centering
    \includegraphics[width=\linewidth]{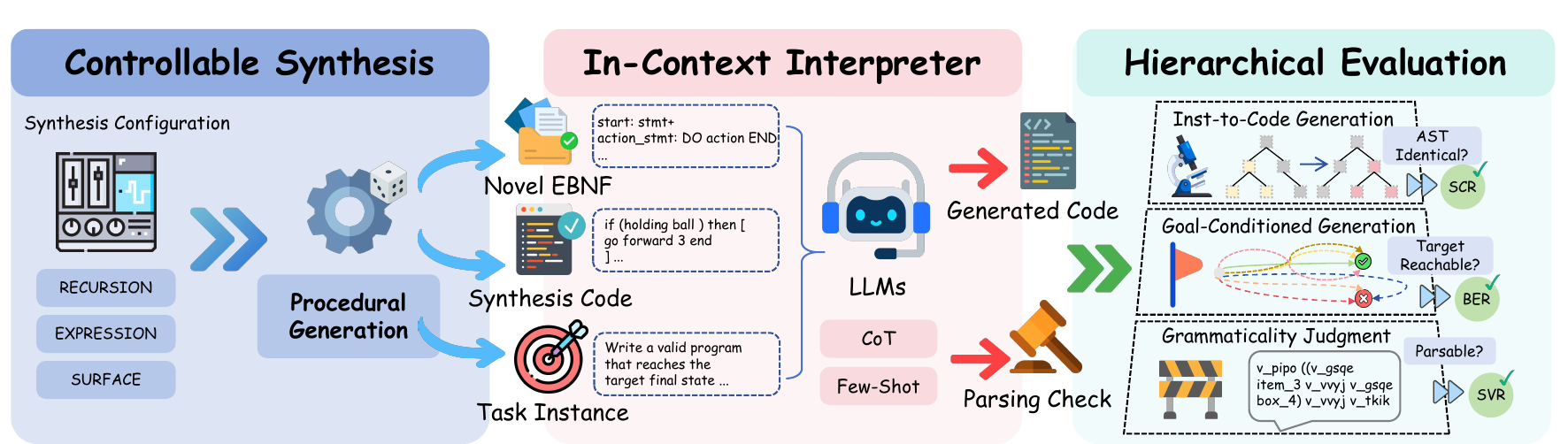}
    \caption{Architecture of the evaluation framework. It illustrates the transition from controllable synthesis of novel grammars to the hierarchical assessment of LLMs' in-context interpretation capability.}
    \label{fig:pipeline}
    \vspace{-5pt}
\end{figure*}
\section{Introduction}
Large language models (LLMs) are increasingly deployed as autonomous agents that interact with their environment through tool-calling~\citep{yao2022react}, API orchestration~\citep{hou2025model}, and the generation of domain-specific languages~\citep{mazrouei2025anka}. In these agentic workflows, the LLM's output must conform to rigid, machine-interpretable interfaces~\citep{shorten2024structuredrag}, such as JSON schemas or function signatures defined dynamically within the prompt~\citep{geng2025jsonschemabench}. This practical requirement implicitly asks LLMs to function as \textbf{in-context grammar interpreters}: given a novel context-free grammar~(CFG)~\citep{chomsky1959algebraic} specification at inference time, can a LLM produce outputs that are syntactically valid, behaviorally functional, and semantically faithful?

While LLMs show apparent fluency in common languages like Python~\citep{jain2024livecodebench}, much of this competence may be attributable to extensive pre-training exposure~\citep{liang2025swe}. This raises a fundamental challenge for agentic reliability: \textit{do LLMs truly internalize the formal logic of a provided grammar, or are they merely retrieving surface regularities from memory?} In real-world tool-use, "almost correct" is insufficient~\citep{yehudai2025survey}. A single missing delimiter or a logical misalignment in a nested loop can lead to catastrophic execution failure~\citep{dou2024s}. To build robust agents, we must understand the LLM's capacity for \textbf{systematic generalization}: the ability to induce and apply structural rules from novel, in-context definitions without relying on familiar semantic cues.

To rigorously investigate this, we propose a controlled evaluation hierarchy that disentangles grammar interpretation into three distinct layers:
(1) \textbf{Syntax Validity}: Does the output conform to the CFG production rules?
(2) \textbf{Behavioral Equivalence}: Does the executed program achieve the intended state change?
(3) \textbf{Semantic Correctness}: Is the induced abstract syntax tree (AST) structurally identical to the ground-truth logic?
This decomposition enables a fine-grained diagnosis of failure modes: whether a LLM fails due to syntactic fragility, planning errors, or a reliance on semantic anchors.

We operationalize this hierarchy through three tasks: grammaticality judgment, goal-conditioned generation, and instruction-to-code generation, conducted within \textsc{RoboGrid}. \textsc{RoboGrid} is a deterministic grid-world environment designed to provide unambiguous execution semantics and a clear mapping between code and behavior, following recent trends in utilizing synthetic environments for precise behavioral evaluation~\citep{yang2025evaluating}. Our framework supports systematic stress-testing across recursion depth, expression complexity, and surface realizations. Notably, we introduce \textbf{"\texttt{Alien}" lexicons}, which replace familiar keywords with opaque tokens, to force the LLM to rely exclusively on the provided grammar, thereby isolating pure symbolic reasoning from semantic priors.

Through systematic experiments, we find that LLMs exhibit a consistent hierarchical degradation in grammar interpretation, where syntactic validity does not guarantee behavioral or semantic correctness under complex constraints. Specifically, even the strongest LLMs struggle with structural alignment, showing a significant gap between surface parsing and logical execution. Furthermore, our ablations reveal that performance is highly sensitive to recursion depth and lexical familiarity. Sharp non-linear decay occurs as depth increases, consistent with findings on the limitations of LLMs in maintaining deep reasoning chains~\citep{rodkin2025beyond}. The sharp decline in performance when using \texttt{Alien} tokens further indicates that LLMs struggle with pure symbolic induction once stripped of familiar lexical priors.

In summary, our contributions are as follows:
\begin{itemize}
\item We develop \textsc{RoboGrid}, a diagnostic benchmark that utilizes a three-layer evaluation hierarchy (\textbf{Syntax, Behavior, Semantics}) to localize failures in structural reasoning.
\item We formalize the task of \textbf{in-context grammar interpretation} for LLMs, shifting the evaluation focus from surface pattern matching to formal rule induction.
\item We characterize the operational limits of SOTA LLMs, revealing a \textbf{massive gap between syntactic adherence and semantic alignment}. While LLMs may follow formal rules at the surface level, their ability to maintain logical fidelity collapses as complexity increases, exposing a heavy reliance on semantic priors rather than pure symbolic induction.
\end{itemize}

\begin{table*}[t]
\centering
\small
\renewcommand{\arraystretch}{1.25}
\setlength{\tabcolsep}{0pt} 
\begin{tabular*}{\textwidth}{@{\extracolsep{\fill}}llcl@{}}
\toprule
\textbf{Category} & \textbf{Parameter} & \textbf{Conf. Space} & \textbf{Evaluation Purpose} \\
\midrule
\multirow{2}{*}{\textsc{Recursion}} & Max Nesting Depth ($D$) & $\{2, \dots, 20\}$ & Probes hierarchical state tracking horizon \\
& Else-Branch Prob. ($p$) & $[0.0,1.0]$ & Modulates control flow branching density \\
\midrule
\textsc{Expression} & Expression Depth ($E$) & $\{1,2,3\}$ & Controls nested arithmetic and boolean predicates \\
\midrule
\multirow{2}{*}{\textsc{Surface}} & Syntactic Style & \{\texttt{Block}, \texttt{C-style}, \texttt{S-expr}\} & Tests robustness to structural delimiters \\
& Lexical Familiarity & \{\texttt{Natural}, \texttt{Alien}\} & Isolates syntactic reasoning from semantic priors \\
\bottomrule
\end{tabular*}
\caption{Controllable parameters for procedural data generation. By systematically varying these dimensions ($D, p, E$), we isolate specific aspects of grammar interpretation capability.}
\label{tab:params}
\vspace{-15pt}
\end{table*}

\section{Preliminaries}
\label{sec:preliminaries}

\subsection{\textsc{RoboGrid} Environment}
\label{sec:robogrid_env}
\textsc{RoboGrid} is a deterministic grid-world designed to decouple a LLM's understanding of task logic from its adherence to formal syntax. In this environment, a robot navigates the grid to pick up objects which are then stored in its inventory. The robot maintains a state $(\texttt{x}, \texttt{y}, \texttt{facing}, \texttt{inventory})$, representing its 2D coordinates, orientation, and the collection of held items respectively. The instruction set is as follows:
\begin{lstlisting}[style=rules]
Move(dir, n)           #forward/backward n steps
Turn(dir)              #left/right 90 deg
Grab(item)             #pick up item|key|box
Loop(n, body)          #repeat body n times
If(cond, then, else?)  #conditional branch
Holding(item)          #state-query predicates
And | Or | Not         #logical operators
\end{lstlisting}

\subsection{Grammar Specification via EBNF}
To interact with \textsc{RoboGrid}, LLMs must generate code conforming to a specific CFG provided in the prompt. We specify these grammars using \textbf{Extended Backus-Naur Form (EBNF)}~\citep{mccracken2003backus}. A representative EBNF snippet for \textsc{RoboGrid} is shown below:

\noindent
\begin{minipage}{\linewidth}
\begin{lstlisting}[style=ebnf]
start: stmt+
stmt: action_stmt | loop | if_stmt

action_stmt: DO action END
loop: LOOP INT TIMES LBR stmt+ RBR
if_stmt: IF cond THEN LBR stmt+ RBR (ELSE LBR stmt+ RBR)?

action: MOVE MOVE_DIR INT? | TURN TURN_DIR | GRAB ITEM
cond: HOLDING ITEM

DO: "exec"      END: "end"
LOOP: "loop"    TIMES: "times"
IF: "when"      THEN: "after"
LBR: "["        RBR: "]"
\end{lstlisting}
\end{minipage}

\noindent This EBNF defines the legal hierarchical structure of the language. Specifically, it specifies:
\begin{itemize}
    \item \textbf{Recursive Composition}: The \texttt{stmt} non-terminal expands into actions, loops, or conditionals, where \texttt{loop} and \texttt{if\_stmt} can recursively contain \texttt{stmt+} to enable nesting.
    \item \textbf{Explicit Delimiters}: Actions are wrapped by \texttt{DO}/\texttt{END} and control blocks by \texttt{LBR}/\texttt{RBR}, providing anchors for tracking recursion levels.
    \item \textbf{Dynamic Terminals}: Bottom-level mappings (e.g., \texttt{LOOP: "loop"}) define the lexicon.
\end{itemize}

By substituting terminals, e.g., replacing \texttt{"loop"} with an opaque token like \texttt{"v\_xkqm"}, we decouple pure syntactic induction from semantic priors. This formalizes LLMs as \textbf{in-context interpreters}, where the EBNF acts as a dynamic protocol that LLMs must interpret on the fly.

\section{Controllable Data Generation}
\label{sec:data_generation}
To rigorously evaluate LLMs' ability to induce novel CFGs, we developed a controllable generation pipeline that disentangles structural complexity from surface representation. Unlike static benchmarks, our framework dynamically synthesizes instances by sweeping across a multi-dimensional configuration space (see Table~\ref{tab:params}).

\subsection{Structural Complexity: Recursion and Expressions}
The core difficulty of a CFG language lies in its hierarchical depth. We control this via two primary dimensions.

\paragraph{Hierarchical Depth} The \textit{Max Nesting Depth} ($D$) limits the recursion level of control structures such as \texttt{Loop} and \texttt{If}. This parameter probes the model's ability to maintain a deep stack of nested execution states. We further modulate control-flow density using \textit{Else-Branch Probability}, which determines the structural branching factor of the program.

\paragraph{Expression Complexity} To decouple high-level logic from local computation, we parameterize the \textit{Expression Depth} ($E$), which controls the nesting level of arithmetic and boolean predicates. This ensures that a LLM's failure in deep recursion can be distinguished from its inability to resolve complex nested logic. By varying $E$, we isolate the impact of local predicate resolution on the overall program's behavioral and semantic correctness.

\subsection{Surface Realization: Styles and Lexicons}
\label{sec:surface-variation}
To isolate symbolic reasoning from the effects of pre-training exposure, we vary the surface realization of program through two key mechanisms.

\paragraph{Syntactic Style} We implement three distinct formatting conventions: \texttt{Block} (verbose keywords like \texttt{do/end}), \texttt{C-style} (braces \texttt{\{\}} and semicolons), and \texttt{S-expr} (Lisp-like prefix notation). This variation tests whether LLMs are biased toward specific structural delimiters inherited from common programming languages.

\paragraph{Lexical Familiarity} This is a critical control in our framework. While \texttt{Natural} mode employs standard keywords (e.g., \texttt{loop}), \texttt{Alien} mode maps all grammar terminals to randomized, opaque tokens (e.g., \texttt{v\_xkqm}). This \texttt{Alien} setting necessitates pure in-context induction, as the LLM cannot leverage semantic priors to infer the function of a command, forcing a reliance on the provided EBNF specification.

\subsection{Synthesis Procedure}
Algorithm~\ref{alg:synthesis} formalizes the generation process. The pipeline first samples a grammar variant $\mathcal{G}$ by binding a syntactic style to a lexicon. It then performs a constrained recursive walk to construct a valid AST within the specified depth limits. Finally, the AST is linearized into a code string $y$ according to the chosen style and mapping. This ensures that every generated instance is guaranteed to be syntactically valid and possesses a deterministic execution trace.

{\setlength{\textfloatsep}{5pt}
\begin{algorithm}[t]
\caption{Controllable Grammar and Program Generation}
\label{alg:synthesis}
\small
\begin{algorithmic}[1]
\Require Styles $\mathcal{S}$, Lexicons $\mathcal{L}$, Max depth $D_{\max}$, Max block size $B_{\max}$, Else-prob $p_{e}$
\Ensure Grammar $\mathcal{G}$, Code $y$, AST $T$
\Statex
\Function{GenerateInstance}{}
    \State $s \sim \mathcal{S}, \ell \sim \mathcal{L}$ \Comment{Sample style and lexicon}
    \State $\mathcal{K} \gets \Call{MapLexicon}{\ell}$ 
    \State $\mathcal{G} \gets \Call{BuildGrammar}{s, \mathcal{K}}$
    \State $T \gets \Call{SampleNode}{d=0}$
    \State $y \gets \Call{Linearize}{T, s, \mathcal{K}}$
    \State \Return $(\mathcal{G}, y, T)$
\EndFunction
\Statex
\Function{SampleNode}{$d$}
    \If{$d \ge D_{\max}$} 
        \State \Return \Call{SampleAction}{} \Comment{Base case: leaf node}
    \EndIf
    \State $t \sim \text{Categorical}(\{\text{Act, Loop, If}\})$
    \If{$t = \text{Act}$}
        \State \Return \Call{SampleAction}{}
    \ElsIf{$t = \text{Loop}$}
        \State $e \gets \Call{SampleExpr}{d}$
        \State $B \gets \Call{SampleBlock}{d+1}$
        \State \Return \textbf{Node}(\texttt{Loop}, $e, B$)
    \ElsIf{$t = \text{If}$}
        \State $c \gets \Call{SampleCond}{d}$
        \State $T_{b} \gets \Call{SampleBlock}{d+1}$
        \State $E_{b} \gets \text{Bernoulli}(p_{e}) ? \Call{SampleBlock}{d+1} : \text{None}$
        \State \Return \textbf{Node}(\texttt{If}, $c, T_{b}, E_{b}$)
    \EndIf
\EndFunction
\Statex
\Function{SampleBlock}{$d$}
    \State $n \sim \text{Uniform}(1, B_{\max})$ \Comment{Number of statements}
    \State \Return $[\Call{SampleNode}{d} \text{ for } i = 1 \dots n]$
\EndFunction
\end{algorithmic}
\end{algorithm}
}

\section{Evaluation Framework} 
\label{sec:framework}

To rigorously assess LLMs as in-context grammar interpreters, we propose a diagnostic framework that disentangles the multi-faceted process of grammar induction. Our framework combines three progressively difficult tasks with a three-layered hierarchy of evaluation metrics.

\subsection{Task Definitions} 
\label{sec:task_defs} 
We design three tasks to evaluate the LLM's ability to recognize, plan, and implement logic under a novel grammar $\mathcal{G}$ provided in the prompt. Detailed examples of each task are provided in Appendix~\ref{sec:appendix-prompts}.

\paragraph{Task 1: Grammaticality Judgment} 
This task probes the LLM's capacity for \textit{syntactic recognition}. Given $\mathcal{G}$ and a candidate string $s$, the LLM must classify it as \texttt{VALID} or \texttt{INVALID}. We construct the test set with a balanced distribution of valid programs and strings containing various perturbations (e.g., delimiter mismatches, illegal nesting, or lexical violations). This task isolates pure syntactic induction from semantic-related reasoning.

\paragraph{Task 2: Goal-Conditioned Generation} 
This task evaluates \textit{behavioral planning} under syntactic constraints. Given $\mathcal{G}$ and a target environment state $g$ (e.g., the robot's final coordinates and inventory), the LLM must synthesize a program $\hat{y}$ that reaches $g$. This assesses whether the LLM can map abstract goals to the provided EBNF primitives, even if the synthesized program structure differs from ground-truth solution.

\paragraph{Task 3: Instruction-to-Code Generation} 
The most demanding task, requiring \textit{procedural alignment}. Given $\mathcal{G}$ and a natural language instruction $x$ (e.g., "Move forward 3 steps, then repeat the grab action twice"), the LLM must generate code $\hat{y}$ that faithfully implements the specified logic. To ensure a precise mapping between instructions and ground-truth code, these natural language descriptions are deterministically synthesized using templates corresponding to the underlying AST structure. This requires the LLM to map the hierarchical structure of natural language to the nested syntax of $\mathcal{G}$ while maintaining semantic fidelity.

\subsection{Evaluation Metrics} 
\label{sec:metrics} 
We evaluate performance across the tasks using a hierarchy of three metrics. 

\paragraph{Syntax Validity Rate (SVR)} 
SVR measures the proportion of LLM's outputs that conform to the production rules of $\mathcal{G}$. For Task 1, SVR is the classification accuracy of the judgment. For Tasks 2 and 3, it is defined as the parseable rate:
\begin{equation} 
\text{SVR} = \mathbb{E} [\mathbf{1}\{\text{parse}(\hat{y}, \mathcal{G}) \neq \text{error}\}]. 
\end{equation}
where $\hat{y}$ denotes the program string synthesized by LLM. The $\text{parse}(\cdot)$ function is implemented via \texttt{Lark} library, which provides a standard parser to validate $\hat{y}$ against $\mathcal{G}$.

\paragraph{Behavioral Equivalence Rate (BER)} 
Applied to Tasks 2 and 3, BER measures the ratio of generated programs that, when executed from an initial state $s_0$, result in a final environment state identical to the ground truth: 
\begin{equation} 
\text{BER} = \mathbb{E} [\mathbf{1}\{\text{exec}(\hat{y}, s_0) = \text{exec}(y^*, s_0)\}].
\end{equation} 
Here, $y^*$ represents the ground-truth program, and $\text{exec}(\cdot, s_0)$ denotes a deterministic execution function that maps a program and an initial environment state $s_0$ to a final state. This metric captures functional correctness, rewarding any valid program $\hat{y}$ that achieves the intended outcome regardless of its internal implementation.

\paragraph{Semantic Correctness Rate (SCR)} 
The strictest metric, primarily utilized for Task 3. SCR measures whether the induced AST of the generated output is structurally identical to the ground-truth AST: 
\begin{equation} 
\text{SCR} = \mathbb{E} [\mathbf{1}\{\text{AST}(\hat{y}) = \text{AST}(y^*)\}].
\end{equation} 
SCR ensures that a LLM has correctly internalized the hierarchical logic and structural intent defined by the grammar, precluding "semantic shortcuts" where a LLM reaches the correct state through an incorrect execution path.

By construction, these metrics follow a strict containment: $\text{SCR} \le \text{BER} \le \text{SVR}$. This allows us to localize exactly where the LLM's reasoning fails---whether at the level of surface syntax, functional execution, or structural semantics.

We further define the conditional metrics as $\text{CBER} = \text{BER} / \text{SVR}$ and $\text{CSCR} = \text{SCR} / \text{SVR}$. These scores isolate the LLM's logical and structural performance from its basic syntactic proficiency, revealing its reasoning quality specifically on validly formatted outputs.

\begin{table*}[t]
\centering
\small
\setlength{\tabcolsep}{5pt} 
\renewcommand{\arraystretch}{1.3}

\setlength{\aboverulesep}{0pt}
\setlength{\belowrulesep}{0pt}

\begin{tabular}{l|c|cc:c|ccc:cc}
\toprule
\multirow{2}{*}{\textbf{Model}} & \textbf{Grammaticality} & \multicolumn{3}{c|}{\textbf{Goal-Conditioned}} & \multicolumn{5}{c}{\textbf{Instruction-to-Code}} \\ 
& \textbf{Judgment} & \multicolumn{3}{c|}{\textbf{Generation}} & \multicolumn{5}{c}{\textbf{Generation}} \\ \cline{2-10}
& \textbf{SVR} & \textbf{SVR} & \textbf{BER} & \textbf{CBER} & \textbf{SVR} & \textbf{BER} & \textbf{SCR} & \textbf{CBER} & \textbf{CSCR} \\
\midrule
\rowcolor{blue!5} \multicolumn{10}{l}{\textit{Open Source Models}} \\
Qwen3-8B      & 47.0 & 1.00 & 1.00 & \textbf{100} & 0.00 & 0.00 & 0.00 & -- & -- \\
Qwen3-32B     & 51.0 & 62.5 & 58.5 & 93.6 & 0.50 & 0.50 & 0.00 & \textbf{100} & 0.00 \\
Qwen3-235B    & \textbf{92.0} & 96.0 & 76.5 & 79.7 & 40.0 & 34.5 & 19.0 & 86.3 & 47.5 \\
Mimo-V2-flash & 79.5 & 69.5 & 54.0 & 77.7 & 21.5 & 17.5 & 9.00 & 81.4 & 41.9 \\
DeepSeek-V3.2 & 86.0 & \textbf{100} & \textbf{96.5} & 96.5 & \textbf{68.5} & 60.0 & \textbf{39.5} & 87.6 & \textbf{57.7} \\
GLM4.7        & 65.0 & 74.5 & 68.5 & 91.9 & 44.5 & 38.5 & 24.5 & 86.5 & 55.1 \\
MiniMax M2.1  & 41.0 & 91.5 & 85.5 & 93.4 & 8.50 & 6.00 & 2.50 & 70.6 & 29.4 \\
\midrule
\rowcolor{yellow!5} \multicolumn{10}{l}{\textit{Closed Source Models}} \\
GPT-5-nano    & 70.0 & 76.5 & 71.5 & 93.5 & 9.50 & 1.50 & 0.00 & 15.8 & 0.00 \\
GPT-5-mini    & 83.5 & \textbf{100} & 90.0 & 90.0 & 65.0 & \textbf{60.5} & 6.00 & 93.1 & 9.23 \\
GPT-5.2       & 75.0 & 95.5 & 60.0 & 62.8 & 38.0 & 16.0 & 10.0 & 42.1 & 26.3 \\
\bottomrule
\end{tabular}
\caption{Main results under the \texttt{Alien} Lexicon setting at Depth 10. All metrics are reported in percentage (\%). Bold indicates the best performance per column.}
\label{tab:main_results}
\vspace{-15pt}
\end{table*}

\section{Experimental Results and Findings}
\label{sec:experiments}
We conduct a comprehensive evaluation of LLMs on \textsc{RoboGrid} to assess their ability to handle recursive syntax and semantic constraints.

\subsection{Main Experiments}
\label{sec:main_experiments}

We evaluate LLMs across three progressively difficult tasks: \textbf{Grammaticality Judgment}, which tests syntactic recognition; \textbf{Goal-Conditioned Generation}, which assesses behavioral planning under constraints; and \textbf{Instruction-to-Code Generation}, which requires full alignment across syntax, behavior, and semantics.

\paragraph{Configurations.}
We establish a high-difficulty benchmark with 200 distinct samples per task, fixing a recursion depth of $D=10$ and a 0-shot Chain of Thought~(CoT)~\citep{wei2022chain} prompting strategy. To isolate symbolic reasoning from semantic memorization, we utilize the \texttt{block} syntactic style paired with an \texttt{Alien} lexicon, where all functional keywords are replaced by opaque tokens. We evaluate a diverse suite of LLMs, including the \text{Qwen3} family (8B, 32B, 235B)~\citep{yang2025qwen3}, reasoning LLMs (\text{DeepSeek-V3.2}~\citep{liu2025deepseek}, \text{Mimo-V2-flash}~\citep{mimo2025flash}, \text{GLM4.7}~\citep{GLM}, \text{MiniMax M2.1})~\citep{MiniMax}, and the \text{GPT-5} series~\citep{gpt5}.

\paragraph{Results and Analysis.}
As reported in Table~\ref{tab:main_results}, performance consistently follows a hierarchical degradation of $\text{SCR} < \text{BER} < \text{SVR}$, highlighting the challenge of maintaining structural fidelity under deep recursion. Three prominent findings emerge: (1) \textbf{Model Scale vs. Task Consistency:} Increased model scale does not consistently translate to better structural alignment. For instance, \text{GPT-5-mini} significantly outperforms the larger \text{GPT-5.2} in Task 2 BER (90.0\% vs. 60.0\%). (2) \textbf{Syntactic Mastery vs. Logical Fidelity:} The conditional metrics CBER and CSCR reveal that syntax is a necessary but insufficient condition for correctness. Notably, \text{DeepSeek-V3.2} maintains a high Task 3 CBER (87.6\%), whereas \text{GPT-5-mini} exhibits a severe semantic collapse with a CSCR of only 9.23\%  despite its high SVR. (3) \textbf{Functional Heuristics vs. Structural Alignment:} The sharp drop in CSCR across all models under the \texttt{Alien} lexicon suggests that without semantic anchors, LLMs rely on surface-level workarounds rather than pure symbolic state-tracking, failing to maintain a consistent mapping between nested instructions and recursive production rules.

\begin{table}[t]
\centering
\small
\renewcommand{\arraystretch}{1.2}
\setlength{\tabcolsep}{2pt}
\begin{tabular}{lp{2.8cm}ccc}
\toprule
\textbf{Dimension} & \textbf{Configuration} & \textbf{SVR} & \textbf{BER} & \textbf{SCR} \\
\midrule
\rowcolor{gray!10} \textbf{Baseline} & \textbf{Standard} & \textbf{21.5} & \textbf{17.5} & \textbf{9.00} \\
\midrule
\multirow{2}{*}{\shortstack[l]{Control Flow\\Density ($p$)}} & No Else ($p=0.0$) & 24.5 & 22.5 & 15.0 \\
& All Else ($p=1.0$) & 13.0 & 11.5 & 8.50 \\
\midrule
\multirow{2}{*}{\shortstack[l]{Expression\\Complexity}} & Shallow ($E=1$) & 22.5 & 20.0 & 10.5 \\
& Deep ($E=3$) & 16.5 & 14.0 & 4.00 \\
\midrule
\multirow{2}{*}{Syntactic Style} & \texttt{C-style} & 19.5 & 16.0 & 6.50 \\
& \texttt{S-expr} & 16.0 & 13.5 & 4.50 \\
\midrule
Lexical Fam. & \texttt{Natural} & 24.5 & 21.5 & 10.5 \\
\bottomrule
\end{tabular}
\caption{Ablation results on \text{Mimo-V2-flash} ($D=10$, 0-shot CoT). All metrics are in \%. $E$ denotes Expression Depth and $p$ denotes Else-branch probability. The \textbf{Standard} baseline uses \texttt{Block} style with \texttt{Alien} lexicon, $E=2$, and $p=0.5$.}
\label{tab:ablation_results}
\vspace{-17pt}
\end{table}

\begin{figure}[t]
    \centering
    \includegraphics[width=1\linewidth]{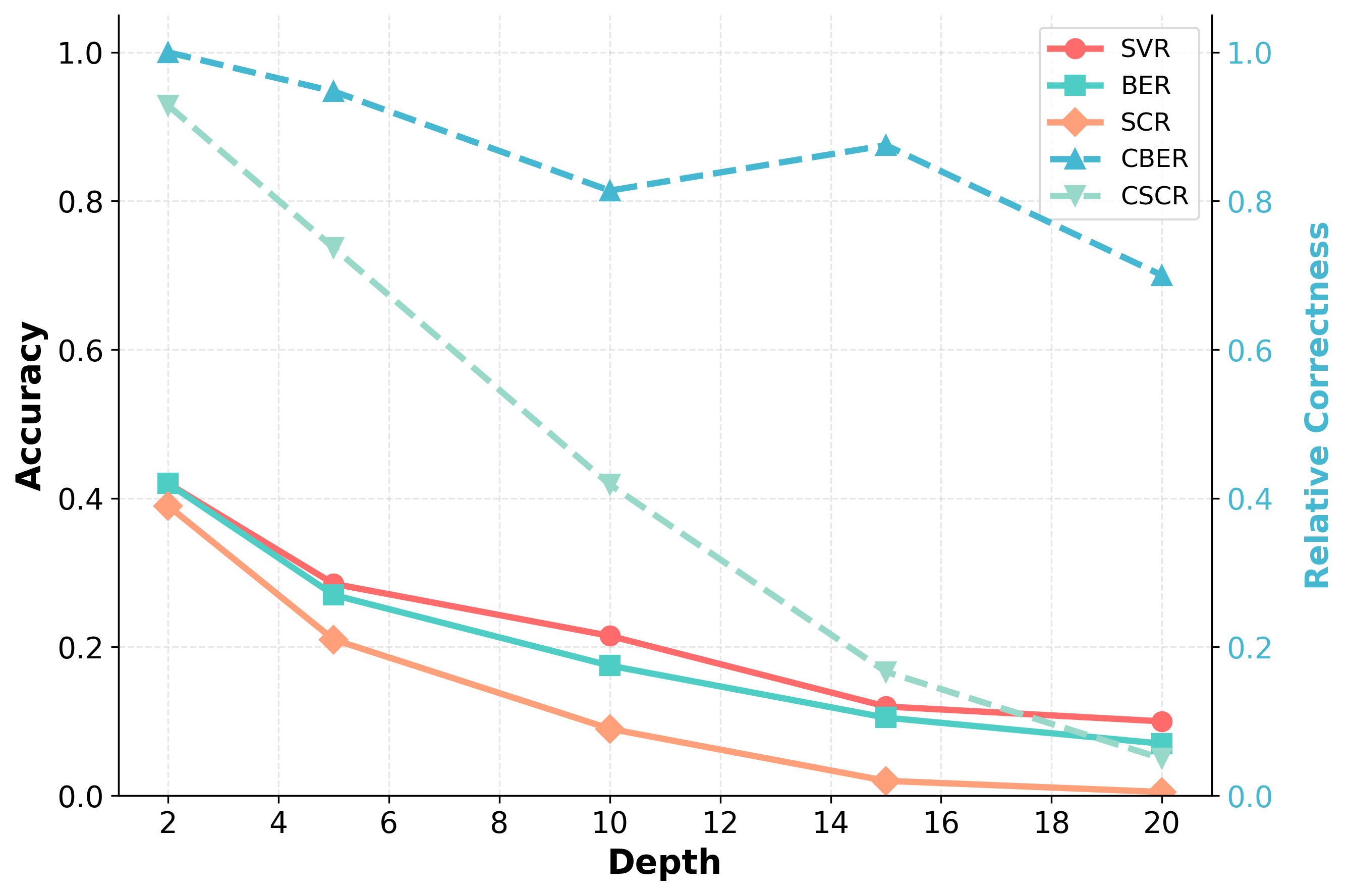}
    \caption{Impact of recursion depth on model performance under the \texttt{Alien} lexicon.}
    \label{fig:depth_nonsense}
    \vspace{-10pt}
\end{figure}

\subsection{Ablation Studies}
\label{sec:ablation}
To further isolate the factors governing in-context grammar induction, we conduct a series of systematic ablations across the three dimensions defined in Table~\ref{tab:params}. For these experiments, we focus on the \textit{Instruction-to-Code Generation} task as our primary testbed, as it requires the most comprehensive alignment across syntax, behavior, and semantics. We employ \text{Mimo-V2-flash} as the representative model for ablation experiments.

\paragraph{Hierarchical Depth and Recursive Scaling} 
The recursive nature of CFGs poses a significant challenge for hierarchical state-tracking. By evaluating LLMs across \textit{Max Nesting Depths} ($D \in \{2, 5, 10, 15, 20\}$), we observe a sharp, non-linear decay across all metrics (Figure~\ref{fig:depth_nonsense}). At $D=2$, the model achieves a moderate SVR of 42.0\% and an SCR of 39.0\%, indicating relatively stable structural alignment at shallow levels. However, as the depth increases to 20, SVR drops to 10.0\%, while SCR nearly vanishes at 0.50\%. This widening gap between syntax and semantics, where the CSCR falls from 92.8\% to a negligible 5.0\%, suggests that while LLMs can occasionally produce locally parseable code at high depths, they almost entirely lose the global state-tracking required to faithfully implement the intended hierarchical logic. 

This structural collapse is further exacerbated by the \textit{Else-Branch Probability} ($p$). As shown in Table~\ref{tab:ablation_results}, forcing $p=1.0$ causes SVR to plunge to 13.0\%, a nearly 40\% reduction compared to the baseline. In contrast, removing else-branches ($p=0.0$) restores SCR to 15.0\%. This confirms that tracking multiple execution paths significantly exhausts the LLM's capacity to maintain the current context. As the density of these branches increases, the model often fails to track the active scope, resulting in incomplete control blocks or the premature termination of nested structures.

\begin{figure}[t]
    \centering
    \includegraphics[width=1\linewidth]{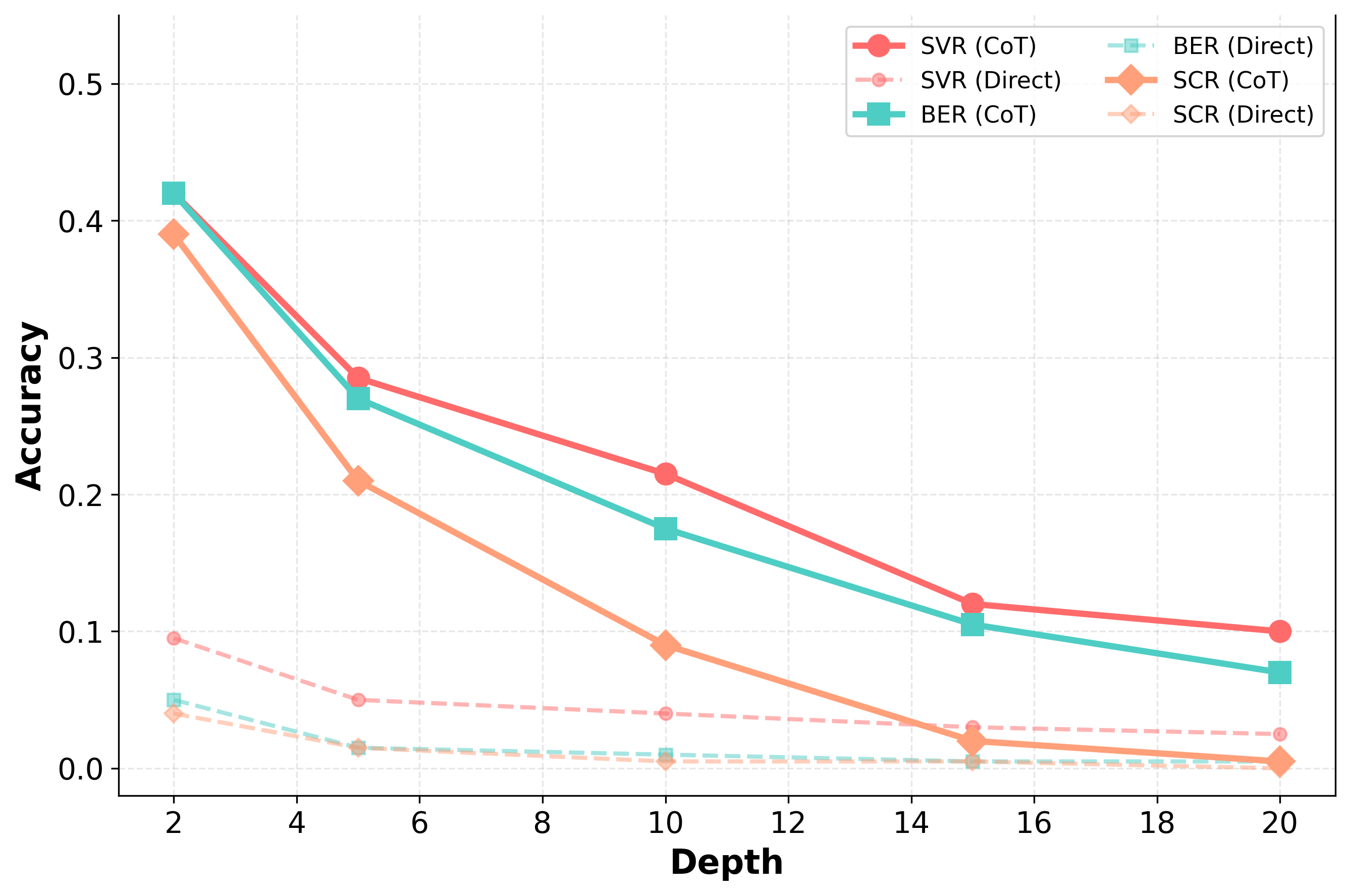}
    \caption{Comparison of CoT and Direct generation across recursion depths.}
    \label{fig:cot_depth_ablation}
    \vspace{-10pt}
\end{figure}

\paragraph{Predicate Complexity and Logic Resolution} 
Beyond the global structure, the complexity of nested predicates within control flow serves as a critical bottleneck. By modulating the \textit{Expression Depth} ($E$), we find that BER and SCR are more sensitive than SVR. As shown in Table~\ref{tab:ablation_results}, increasing the complexity from the baseline ($E=2$) to the deep setting ($E=3$) causes SCR to drop by more than half, from 9.00\% to 4.00\%. Conversely, simplifying predicates to $E=1$ improves SCR to 10.5\% and BER to 20.0\%. The fact that SVR remains relatively high (16.5\%) even at $E=3$ indicates that while LLMs can still generate syntactically valid shells, they fail to resolve the underlying logical operations. This suggests that nested arithmetic and boolean predicates constitute a distinct reasoning hurdle that persists even when the model successfully adheres to the global grammar rules.

\paragraph{Structural Delimiters and Lexical Priors} 
The impact of surface realization reveals the extent to which LLMs rely on pre-training biases versus formal rules. As shown in Table \ref{tab:ablation_results}, among the tested \textit{Syntactic Styles}, the \texttt{Block} used in our baseline is the most robust, while \texttt{C-style} and \texttt{S-expr} lead to consistent performance drops. Specifically, switching to \texttt{S-expr} reduces SVR from 21.5\% to 16.0\% and nearly halves the SCR to 4.50\%. This suggests that without familiar semantic cues, the repetitive parentheses of \texttt{S-expr} become harder for the model to track than explicit keywords. A more profound effect is observed through \textit{Lexical Familiarity}. Moving from the \texttt{Alien} baseline to the \texttt{Natural} lexicon improves SVR to 24.5\% and BER to 21.5\%. This performance gain confirms that LLMs employ semantic bootstrapping,  leveraging familiar keywords like \texttt{loop} to guess behavior, rather than performing pure in-context induction from the provided EBNF specification.

\begin{figure}[t]
    \centering
    \includegraphics[width=1\linewidth]{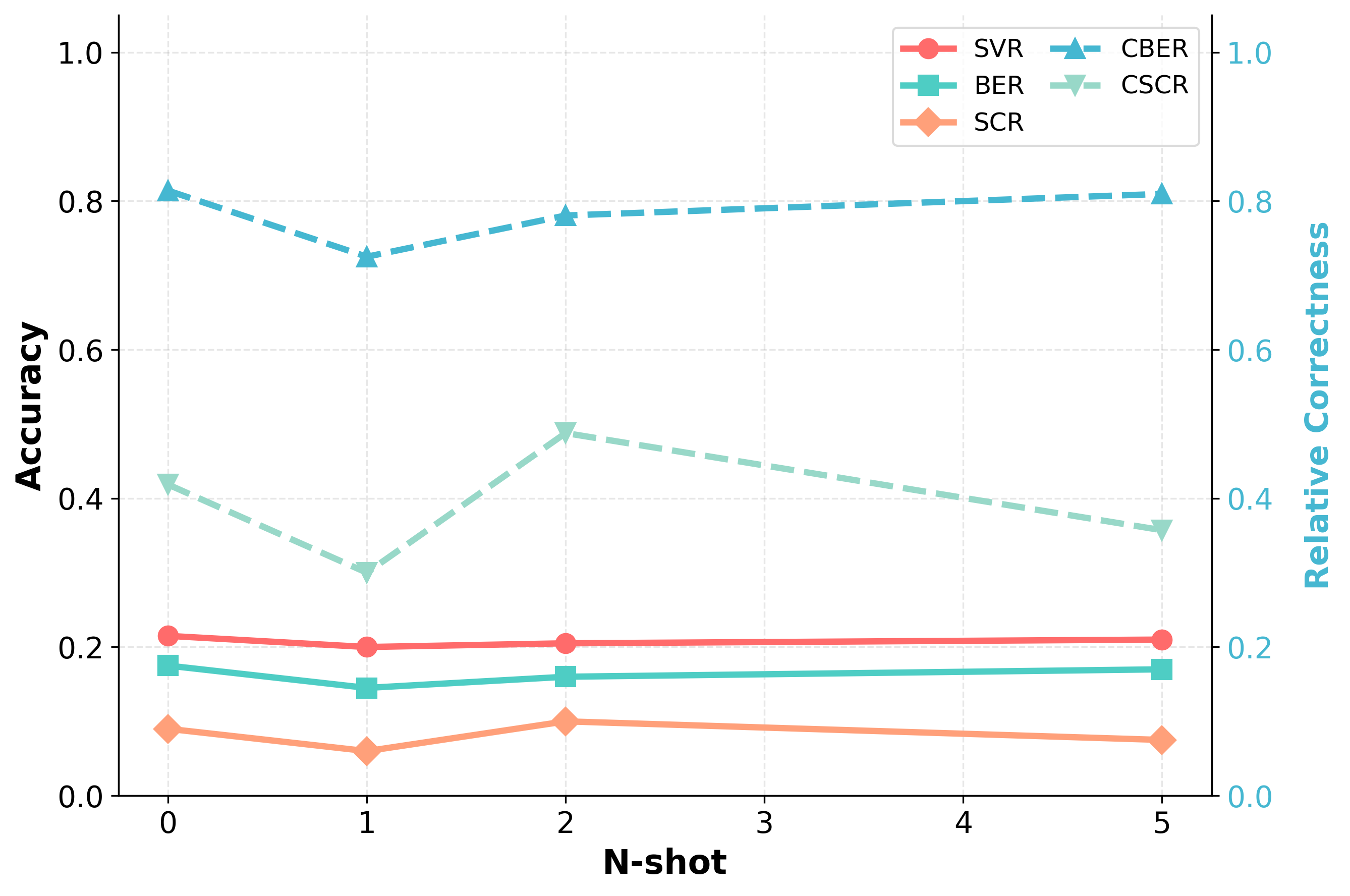}
    \caption{Impact of few-shot examples on performance at fixed recursion depth with CoT reasoning.}
    \label{fig:shot_ablation}
    \vspace{-10pt}
\end{figure}
%==============================================================================
\section{Diagnostic Analysis and Discussion}
\label{sec:analysis}
%==============================================================================

\subsection{Prompting Strategies and ICL Efficiency}
\label{sec:analysis-prompting}
We evaluate the impact of different prompting configurations, specifically comparing zero-shot and few-shot in-context learning~(ICL)~\citep{brown2020language} both with and without CoT reasoning.

\paragraph{The Dominance of CoT in Novel CFGs.} 
Figure~\ref{fig:cot_depth_ablation} reveals that CoT is a critical factor for enabling LLMs to function as grammar interpreters. In the direct~(without CoT) setting, performance is consistently abysmal across all depths, with SVR rarely exceeding 10\% and SCR dropping to near zero even at shallow depths ($D=5$). In stark contrast, enabling CoT provides a massive performance boost, lifting SVR at $D=2$ from 9.5\% to 42.0\% with zero-shot setting. This suggests that the external reasoning trace is a functional necessity for the model to maintain the hierarchical state-tracking required by EBNF production rules.

\paragraph{Diminishing Returns of Few-shot Prompting.} 
To isolate the impact of few-shot examples, we analyze performance variations under the CoT setting specifically at a fixed complexity of $D=10$ (Figure~\ref{fig:shot_ablation}). We observe diminishing returns as the number of shots increases. Moving from 0-shot (21.5\% SVR, 9\% SCR) to 1-shot actually results in a performance dip (20.0\% SVR, 6.0\% SCR). While 2-shot and 5-shot trials show minor recoveries in SVR (20.5\% and 21.0\% respectively), they fail to consistently surpass the 0-shot baseline. This indicates that while ICL might assist with surface-level alignment in simpler tasks, for deep recursive structures, additional examples do not provide a clear reasoning advantage and may even introduce distracting noise into the context window.

\paragraph{Universal Fragility at Depth.} 
Regardless of the prompting strategy, all configurations exhibit a sharp performance decay as recursion depth increases (Figure~\ref{fig:cot_depth_ablation}). Even with CoT, SCR plunges from 39.0\% at $D=2$ to a negligible 0.5\% at $D=20$. This universal downward trend confirms that current LLMs remain fundamentally limited in their ability to maintain deep symbolic abstractions, with neither CoT nor ICL fully mitigating the structural collapse at extreme hierarchical depths.

\subsection{Failure Mode and Case Studies}
\label{sec:analysis-errors}

To diagnose the operational limits of LLMs, we categorize the observed failures into a three-layered taxonomy, analyzing how reasoning chains break down at each level of the evaluation hierarchy. A detailed taxonomy and abstracted case studies for each category are provided in Appendix~\ref{sec:appendix-failure-modes}.

\paragraph{Syntactic Fragility} 
Syntactic failures manifest as a fundamental breakdown in following formal production rules. LLMs frequently suffer from \textbf{lexical mapping collapses}, \textbf{operator confusion}, \textbf{scoping violations}, and \textbf{delimiter imbalances}, exposing LLMs' struggle to maintain structural integrity under strict symbolic constraints.

\paragraph{Behavioral Misalignment} 
Behavioral failures occur when predicted code is syntactically valid but fails to achieve the target. These typically manifest as \textbf{logical predicate inversions} or \textbf{arithmetic simulation collapses} where mathematical expressions are miscalculated. we also observe a \textbf{planning horizon collapse} where LLMs prematurely terminate the execution flow after navigating sub-tasks, leaving the final actions unimplemented.

\paragraph{Semantic Drift}
The most subtle failure mode is semantic drift, where the generated code is behaviorally functional but semantically non-aligned. This is primarily driven by two phenomena: (1) \textbf{AST flattening bias} in arithmetic and boolean contexts, where LLMs prematurely evaluate expressions into simplified results. (2) \textbf{Autoregressive Echoing}, where the model loses its position within deep nested sequences and redundantly duplicates entire logical blocks. These failures reveal that LLMs often prioritize simulative success over structural fidelity, acting as independent executors rather than faithful in-context grammar interpreters.

%==============================================================================
\section{Related Work}
\label{sec:related}
%==============================================================================
\subsection{Computational Expressivity and Grammar Interpretation}
\label{sec:related_1}
While Transformers with polynomial-length CoT theoretically possess the expressivity to recognize CFGs~\citep{merrill2023expressive}, potentially via attention heads simulating dynamic programming~\citep{allen2023physics} or MLP neurons tracking states~\citep{zhang-etal-2025-finite}, a significant gap exists in practice. LLMs often struggle with deep recursion~\citep{schulz2025unraveling} and revert to shallow heuristics over compositional parsing~\citep{petty2025relic}, failing at explicit metalinguistic deduction from formal specifications~\citep{liu2025gold}. These limitations suggest a reliance on statistical pattern matching rather than abstract rule internalization. We extend these insights by using a 3-layered evaluation and \texttt{Alien} lexicons to isolate pure symbolic induction from lexical priors.

\subsection{Formal Constraint Enforcement}
\label{sec:related_2}
To bridge the gap between probabilistic generation and rigid machine interfaces, researchers have developed enforcement frameworks that ensure syntactic validity. Grammar-Constrained Decoding (GCD) engines, such as XGrammar~\citep{dong2025xgrammar}, SynCode~\citep{ugare2024syncode}, and TreeCoder~\citep{princis2025treecoder}, use pushdown automata to mask output distributions, while Grammar-Aligned Decoding~\citep{park2024grammar} aims to guarantee EBNF adherence without semantic distortion. Complementarily, Rule-Following Fine-Tuning (RFFT) methods~\citep{hu2024case, hu2025beyond} attempt to internalize these constraints by shifting model behavior from pattern matching toward deterministic symbolic execution.

\section{Conclusion}
\label{sec:conclusion}

In this work, we formalize the task of in-context grammar interpretation, evaluating LLMs across the three layers of syntax, behavior, and semantics. Through our \textsc{RoboGrid} framework, we demonstrate that while SOTA LLMs can maintain surface syntactic validity, they suffer from a severe syntax-logic gap as structural density increases. Our findings highlight that CoT reasoning is a functional necessity for following formal rules, yet LLMs remain heavily dependent on semantic bootstrapping from familiar keywords. The rapid collapse of semantic alignment under deep recursion and high branching factors pinpoints a fundamental limitation in current hierarchical state-tracking. Ultimately, our results suggest that achieving reliable, grammar-agnostic agents requires moving beyond surface-level prompt engineering toward capabilities of robust symbolic induction and reliable state-tracking across dense hierarchical structures.

%==============================================================================

%==============================================================================
\section*{Limitations}
Despite the insights provided by our evaluation hierarchy, this work has certain limitations that open avenues for future research. 

First, while \textsc{RoboGrid} provides a deterministic and controllable environment for isolating structural reasoning, it relies on template-synthesized instructions. In real-world agentic deployments, LLMs often encounter a complex interleaving of nuanced natural language and highly intricate structured interfaces. The interplay between linguistic ambiguity and formal rigidity may introduce additional failure modes. Future work will extend our framework to more heterogeneous environments to evaluate LLMs as in-context interpreters under open-domain constraints.

Second, although our study exposes a massive gap between syntactic adherence and semantic alignment in novel grammars, we do not propose a specialized architecture or decoding strategy to bridge this divide. Current solutions often either rely on model-internal knowledge or sacrifice semantic flexibility for syntactic validity. Developing a grammar-agnostic reasoning mechanism that maintains both structural integrity and logical fidelity without over-reliance on semantic priors remains a critical challenge for the community.

%==============================================================================

\section*{Ethics Statement}

All experiments in this work are conducted on fully synthetic datasets generated by predefined rules. These datasets do not correspond to real individuals or real-world text and therefore contain no personally identifiable information or offensive content.

During the preparation of this manuscript, AI-assisted tools were used for minor tasks, including 
text polishing and code debugging.

\bibliography{custom}

@article{liu2025gold,
  title={The Gold Medals in an Empty Room: Diagnosing Metalinguistic Reasoning in LLMs with Camlang},
  author={Liu, Fenghua and Chen, Yulong and Liu, Yixuan and Jin, Zhujun and Tsai, Solomon and Zhong, Ming},
  journal={arXiv preprint arXiv:2509.00425},
  year={2025}
}

@article{allen2023physics,
  title={Physics of language models: Part 1, learning hierarchical language structures},
  author={Allen-Zhu, Zeyuan and Li, Yuanzhi},
  journal={arXiv preprint arXiv:2305.13673},
  year={2023}
}

@article{merrill2023expressive,
  title={The expressive power of transformers with chain of thought},
  author={Merrill, William and Sabharwal, Ashish},
  journal={arXiv preprint arXiv:2310.07923},
  year={2023}
}

@article{schulz2025unraveling,
  title={Unraveling Syntax: How Language Models Learn Context-Free Grammars},
  author={Schulz, Laura Ying and Mitropolsky, Daniel and Poggio, Tomaso},
  journal={arXiv preprint arXiv:2510.02524},
  year={2025}
}

@inproceedings{zhang-etal-2025-finite,
    title = "Finite State Automata Inside Transformers with Chain-of-Thought: A Mechanistic Study on State Tracking",
    author = "Zhang, Yifan  and
      Du, Wenyu  and
      Jin, Dongming  and
      Fu, Jie  and
      Jin, Zhi",
    editor = "Che, Wanxiang  and
      Nabende, Joyce  and
      Shutova, Ekaterina  and
      Pilehvar, Mohammad Taher",
    booktitle = "Proceedings of the 63rd Annual Meeting of the Association for Computational Linguistics (Volume 1: Long Papers)",
    month = jul,
    year = "2025",
    address = "Vienna, Austria",
    publisher = "Association for Computational Linguistics",
    url = "https://aclanthology.org/2025.acl-long.668/",
    doi = "10.18653/v1/2025.acl-long.668",
    pages = "13603--13621",
    ISBN = "979-8-89176-251-0",

}

@article{petty2025relic,
  title={RELIC: Evaluating Compositional Instruction Following via Language Recognition},
  author={Petty, Jackson and Hu, Michael Y and Wang, Wentao and Ravfogel, Shauli and Merrill, William and Linzen, Tal},
  journal={arXiv preprint arXiv:2506.05205},
  year={2025}
}

@article{dong2025xgrammar,
  title={Xgrammar: Flexible and efficient structured generation engine for large language models},
  author={Dong, Yixin and Ruan, Charlie F and Cai, Yaxing and Xu, Ziyi and Zhao, Yilong and Lai, Ruihang and Chen, Tianqi},
  journal={Proceedings of Machine Learning and Systems},
  volume={7},
  year={2025}
}

@article{ugare2024syncode,
  title={SynCode: LLM generation with grammar augmentation},
  author={Ugare, Shubham and Suresh, Tarun and Kang, Hangoo and Misailovic, Sasa and Singh, Gagandeep},
  journal={Transactions on Machine Learning Research},
  year={2024}
}

@article{princis2025treecoder,
  title={TreeCoder: Systematic Exploration and Optimisation of Decoding and Constraints for LLM Code Generation},
  author={Princis, Henrijs and Sharma, Arindam and David, Cristina},
  journal={arXiv preprint arXiv:2511.22277},
  year={2025}
}

@article{park2024grammar,
  title={Grammar-aligned decoding},
  author={Park, Kanghee and Wang, Jiayu and Berg-Kirkpatrick, Taylor and Polikarpova, Nadia and D'Antoni, Loris},
  journal={Advances in Neural Information Processing Systems},
  volume={37},
  pages={24547--24568},
  year={2024}
}

@article{hu2025beyond,
  title={Beyond Single-Task: Robust Multi-Task Length Generalization for LLMs},
  author={Hu, Yi and Kang, Shijia and Yang, Haotong and Xu, Haotian and Zhang, Muhan},
  journal={arXiv preprint arXiv:2502.11525},
  year={2025}
}

@inproceedings{hu2024case,
author = {Hu, Yi and Tang, Xiaojuan and Yang, Haotong and Zhang, Muhan},
title = {Case-based or rule-based: how do transformers do the math?},
booktitle = {Proceedings of the 41st International Conference on Machine Learning},
year = {2024},
series = {ICML'24}
}

@article{yang2025qwen3,
  title={Qwen3 technical report},
  author={Yang, An and Li, Anfeng and Yang, Baosong and Zhang, Beichen and Hui, Binyuan and Zheng, Bo and Yu, Bowen and Gao, Chang and Huang, Chengen and Lv, Chenxu and others},
  journal={arXiv preprint arXiv:2505.09388},
  year={2025}
}

@article{liu2025deepseek,
  title={Deepseek-v3. 2: Pushing the frontier of open large language models},
  author={DeepSeek Team},
  journal={arXiv preprint arXiv:2512.02556},
  year={2025}
}

@misc{mimo2025flash,
  title={MiMo-V2-Flash Technical Report},
  author={LLM-Core Xiaomi},
  year={2025},
  url={https://github.com/XiaomiMiMo/MiMo-V2-Flash/paper.pdf}
}

@misc{gpt5,
  title={Introducing GPT-5},
  author={OpenAI},
  year={2025},
  url={https://openai.com/}
}

@misc{GLM,
      title={GLM-4.7}, 
      author={GLM Team},
      year={2025},
      url={https://docs.z.ai/guides/llm/glm-4.7}, 
}

@misc{MiniMax,
      title={MiniMax M2.1}, 
      author={MiniMax Team},
      year={2025},
      url={https://huggingface.co/MiniMaxAI/MiniMax-M2.1}, 
}

@inproceedings{yao2022react,
  title={React: Synergizing reasoning and acting in language models},
  author={Yao, Shunyu and Zhao, Jeffrey and Yu, Dian and Du, Nan and Shafran, Izhak and Narasimhan, Karthik R and Cao, Yuan},
  booktitle={The eleventh international conference on learning representations},
  year={2022}
}

@article{wei2022chain,
  title={Chain-of-thought prompting elicits reasoning in large language models},
  author={Wei, Jason and Wang, Xuezhi and Schuurmans, Dale and Bosma, Maarten and Xia, Fei and Chi, Ed and Le, Quoc V and Zhou, Denny and others},
  journal={Advances in neural information processing systems},
  volume={35},
  pages={24824--24837},
  year={2022}
}

@article{brown2020language,
  title={Language models are few-shot learners},
  author={Brown, Tom and Mann, Benjamin and Ryder, Nick and Subbiah, Melanie and Kaplan, Jared D and Dhariwal, Prafulla and Neelakantan, Arvind and Shyam, Pranav and Sastry, Girish and Askell, Amanda and others},
  journal={Advances in neural information processing systems},
  volume={33},
  pages={1877--1901},
  year={2020}
}

@incollection{chomsky1959algebraic,
  title={The algebraic theory of context-free languages},
  author={Chomsky, Noam and Sch{\"u}tzenberger, Marcel P},
  booktitle={Studies in Logic and the Foundations of Mathematics},
  volume={26},
  pages={118--161},
  year={1959},
  publisher={Elsevier}
}

@article{hou2025model,
  title={Model context protocol (mcp): Landscape, security threats, and future research directions},
  author={Hou, Xinyi and Zhao, Yanjie and Wang, Shenao and Wang, Haoyu},
  journal={arXiv preprint arXiv:2503.23278},
  year={2025}
}

@article{mazrouei2025anka,
  title={Anka: A Domain-Specific Language for Reliable LLM Code Generation},
  author={Mazrouei, Saif Khalfan Saif Al},
  journal={arXiv preprint arXiv:2512.23214},
  year={2025}
}

@article{shorten2024structuredrag,
  title={Structuredrag: Json response formatting with large language models},
  author={Shorten, Connor and Pierse, Charles and Smith, Thomas Benjamin and Cardenas, Erika and Sharma, Akanksha and Trengrove, John and van Luijt, Bob},
  journal={arXiv preprint arXiv:2408.11061},
  year={2024}
}

@article{yehudai2025survey,
  title={Survey on evaluation of llm-based agents},
  author={Yehudai, Asaf and Eden, Lilach and Li, Alan and Uziel, Guy and Zhao, Yilun and Bar-Haim, Roy and Cohan, Arman and Shmueli-Scheuer, Michal},
  journal={arXiv preprint arXiv:2503.16416},
  year={2025}
}

@article{liang2025swe,
  title={The SWE-Bench Illusion: When State-of-the-Art LLMs Remember Instead of Reason},
  author={Liang, Shanchao and Garg, Spandan and Moghaddam, Roshanak Zilouchian},
  journal={arXiv preprint arXiv:2506.12286},
  year={2025}
}

@article{jain2024livecodebench,
  title={Livecodebench: Holistic and contamination free evaluation of large language models for code},
  author={Jain, Naman and Han, King and Gu, Alex and Li, Wen-Ding and Yan, Fanjia and Zhang, Tianjun and Wang, Sida and Solar-Lezama, Armando and Sen, Koushik and Stoica, Ion},
  journal={arXiv preprint arXiv:2403.07974},
  year={2024}
}

@article{geng2025jsonschemabench,
  title={Jsonschemabench: A rigorous benchmark of structured outputs for language models},
  author={Geng, Saibo and Cooper, Hudson and Moskal, Micha{\l} and Jenkins, Samuel and Berman, Julian and Ranchin, Nathan and West, Robert and Horvitz, Eric and Nori, Harsha},
  journal={arXiv preprint arXiv:2501.10868},
  year={2025}
}

@article{dou2024s,
  title={What's wrong with your code generated by large language models? an extensive study},
  author={Dou, Shihan and Jia, Haoxiang and Wu, Shenxi and Zheng, Huiyuan and Zhou, Weikang and Wu, Muling and Chai, Mingxu and Fan, Jessica and Huang, Caishuang and Tao, Yunbo and others},
  journal={arXiv preprint arXiv:2407.06153},
  year={2024}
}

@article{yang2025evaluating,
  title={Evaluating the generalization capabilities of large language models on code reasoning},
  author={Yang, Rem and Dai, Julian and Vasilakis, Nikos and Rinard, Martin},
  journal={arXiv preprint arXiv:2504.05518},
  year={2025}
}

@article{rodkin2025beyond,
  title={Beyond Memorization: Extending Reasoning Depth with Recurrence, Memory and Test-Time Compute Scaling},
  author={Rodkin, Ivan and Orel, Daniil and Smirnov, Konstantin and Bolatov, Arman and Elbouardi, Bilal and Hassan, Besher and Kuratov, Yuri and Bulatov, Aydar and Nakov, Preslav and Baldwin, Timothy and others},
  journal={arXiv preprint arXiv:2508.16745},
  year={2025}
}

@incollection{mccracken2003backus,
  title={Backus-naur form (bnf)},
  author={McCracken, Daniel D and Reilly, Edwin D},
  booktitle={Encyclopedia of computer science},
  publisher={John Wiley \& Sons},
  pages={129--131},
  year={2003}
}

\clearpage
%==============================================================================
\appendix
\section{Appendix}
\label{sec:appendix}
%==============================================================================

\subsection{EBNF Examples}
\label{sec:appendix-ebnf}

\subsubsection{Different Syntactic Styles}
We implement three distinct syntactic styles to evaluate the LLMs' sensitivity to structural delimiters and formatting conventions. 

The \texttt{block} style utilizes explicit, verbose keywords to demarcate action and control boundaries, providing clear anchor points for hierarchical parsing:
\begin{lstlisting}[style=ebnf]
start: stmt+
action_stmt: DO action END
loop: LOOP expr TIMES LBR stmt+ RBR
if_stmt: IF cond THEN LBR stmt+ RBR (ELSE LBR stmt+ RBR)?
...
\end{lstlisting}

The \texttt{C-style} mimics common imperative programming languages, employing curly braces for scoping and semicolons for statement termination:
\begin{lstlisting}[style=ebnf]
start: stmt*
action_stmt: action SEMI
loop: LOOP PAR_L expr PAR_R LBR stmt* RBR
if_stmt: IF PAR_L cond PAR_R LBR stmt* RBR (ELSE LBR stmt* RBR)?
...
\end{lstlisting}

The \texttt{S-expr} style adopts a Lisp-like prefix notation with extensive parenthesization:
\begin{lstlisting}[style=ebnf]
start: stmt+
loop: PAR_L LOOP expr stmt+ PAR_R
if_stmt: PAR_L IF cond THEN stmt+ (ELSE stmt+)? PAR_R
...
\end{lstlisting}

\subsubsection{Different Surface Realization}
To investigate the impact of lexical priors, we vary the mapping of grammar terminals while keeping the underlying production rules constant.

The \texttt{Natural} lexicon employs familiar, semantically meaningful English keywords to represent operations and control flow.
\begin{lstlisting}[style=ebnf]
start: stmt+
action_stmt: DO action END
...
DO: "do";  END: "stop";  
LOOP: "repeat";  TIMES: "times";  
IF: "when";  THEN: "next";
...
\end{lstlisting}
The \texttt{Alien} lexicon replaces these familiar terms with randomized, opaque tokens. This configuration forces the model to rely exclusively on the provided EBNF specification for symbolic induction, as the semantic cues are entirely stripped.
\begin{lstlisting}[style=ebnf]
start: stmt+
action_stmt: DO action END
...
DO: "v_qnuc";  END: "v_ojrp";  
LOOP: "v_takv";  TIMES: "v_gqge";  
IF: "v_abgh";  THEN: "v_xhfm";
...
\end{lstlisting}

\subsection{Task Examples}
\label{sec:appendix-prompts}

In this section, we provide concrete examples for the three tasks.
\paragraph{Task 1: Grammaticality Judgment}
The model is provided with a complete EBNF grammar and a code snippet. It must determine if the code is syntactically valid according to the production rules.

\noindent
\begin{minipage}{\linewidth}
\begin{lstlisting}[style=taskbox]
[PROMPT INPUT]
You are a strict syntax checker
Your task is to determine if the provided code is strictly valid according to the EBNF grammar.
EBNF:
start: stmt+
stmt: action_stmt | loop | if_stmt
action_stmt: "act" action "fin"
loop: "repeat" expr "iters" "{" stmt+ "}"
if_stmt: "when" cond "next" "{" stmt+ "}"
...

Code to Check:
repeat (3 + 4) iters {
  repeat ((1 + 0) * 4) iters {
    ...
  }
}

If the code can be parsed by the grammar, output 'VALID'
\end{lstlisting}
\end{minipage}
\paragraph{Task 2: Goal-Conditioned Generation}
The model receives the EBNF grammar, a start state and a target state description. It must synthesize a valid program that achieves this state.

\begin{lstlisting}[style=taskbox]
You are a strict code generator.
You are given a NEW programming language definition (EBNF).
EBNF:
start: stmt+
stmt: action_stmt | loop | if_stmt
action_stmt: "act" action "fin"
loop: "repeat" expr "iters" "{" stmt+ "}"
if_stmt: "when" cond "next" "{" stmt+ "}"
...

Start state: pos (0, 0), facing N, inventory empty. Target final state: pos (0, 46), facing W, inventory empty.

Your task is to synthesize a valid code according to the EBNF that guides the robot to the target state.
\end{lstlisting}

\paragraph{Task 3: Instruction-to-Code Generation}
The model is given the EBNF grammar and a natural language instruction generated by template. It must generate a program that is both syntactically valid and semantically faithful.

\noindent 
\begin{minipage}{\linewidth}
\begin{lstlisting}[style=taskbox]
[PROMPT INPUT]
You are a strict code generator.
You are given a NEW programming language definition (EBNF).
EBNF:
start: stmt+
stmt: action_stmt | loop | if_stmt
action_stmt: "act" action "fin"
loop: "repeat" expr "iters" "{" stmt+ "}"
if_stmt: "when" cond "next" "{" stmt+ "}"
...

Instructions:
Step 1: Move backward ((4 times 4) times (4 plus 3)) steps. Step 2: Move forward (5 plus 1) steps. Step 3: Repeat ((4 times 0) plus 4) times: [ If (not (holding key)) and ((holding box_0) and (holding cube)), then: [ Turn left. ] Otherwise: [ Move backward 3 steps. ] ] Step 4: If not ((holding box_2) and (holding ball_0)), then: [ If not ((holding box) and (holding key)), then: [ Grab the key_2. ] ] Otherwise: [ Turn right. ]

Your task is to faithfully translate the instructions into code. You must strictly adhere to the provided EBNF syntax and ensure the hierarchical structure of the generated program exactly matches the logical nesting of the instructions.

\end{lstlisting}
\end{minipage}

\subsection{Generated Data Examples}
\label{sec:appendix-examples}

We illustrate the diversity of our generated data with examples covering different recursion depths, syntactic styles, and lexical variations.

\subsubsection{Recursion Depth Variations}

We vary the maximum nesting depth of the AST to probe the LLM's ability to handle hierarchical structures.
At depth 2, programs consist of shallow control structures, typically single loops or conditionals.

\noindent
\begin{minipage}{\linewidth}
\begin{lstlisting}[style=example]
# Low Recursion Depth (Depth 2)
loop (3 + 4) iters {
  loop ((1 + 0) * 4) iters {
    act turn left end
  }
}
loop 3 iters {
  loop (4 * (4 * 2)) iters {
    act turn left end
  }
}
act turn left end
act take box_4 end
if not ((has cube_0 and has cube_4)) after {
  act turn left end
}
\end{lstlisting}
\end{minipage}
At depth 5, programs exhibit moderate nesting, requiring the model to track state across multiple indentations.

\noindent 
\begin{minipage}{\linewidth}
\begin{lstlisting}[style=example]
# Medium Recursion Depth (Depth 5)
if (no (has ball_3) alt (has item and has ball_2)) next {
  if (no (has key_4) alt (has ball_2 alt has ball)) next {
    do go forward (3 * (1 * 1)) end
  }
} else {
  if ((has box and has key_2) and (has ball_0 and has box_0)) next {
    do go backward ((2 + 1) * 4) end
  }
}
loop ((4 * 4) + 3) times {
  do turn left end
}
...
\end{lstlisting}
\end{minipage}

At depth 10, programs feature deep recursion chains, challenging the LLM's ability to maintain hierarchical state-tracking.

\noindent 
\begin{minipage}{\linewidth}
\begin{lstlisting}[style=example]
# High Recursion Depth (Depth 10)
loop ((1 * 1) + (0 + 0)) x [
  exec go forward 5 stop
]
exec go forward 5 stop
loop ((0 * 3) * (3 * 3)) x [
  exec turn left stop
]
loop ((0 + 4) + (0 + 1)) x [
  loop 1 x [
    loop 4 x [
      loop ((2 * 1) * 1) x [
        loop 0 x [
          loop 1 x [
            if ((holding ball_4 alt holding item_1) alt not (holding key_2)) then [
              ...
            ]
          ]
        ]
      ]
    ]
  ]
]
\end{lstlisting}
\end{minipage}

\subsubsection{Expression Complexity Variations}

We also control the complexity of arithmetic and boolean expressions nested within the control flow.
Shallow expressions (depth 1) involve simple arithmetic or boolean operations.

\noindent 
\begin{minipage}{\linewidth}
\begin{lstlisting}[style=example]
# Shallow Expressions (Depth 1)
run (3 + 2) times [
  if (holding ball or holding box) then [
    go forward 3 end
  ]
]
\end{lstlisting}

Deep expressions (depth 3) involve nested operations, requiring correct operator precedence handling.

\begin{lstlisting}[style=example]
# Deep Expressions (Depth 3)
run ((3 * (2 + 1)) + (4 * (0 + 1))) times [
  if ((holding ball and holding box) or (not (holding key) and holding item)) then [
    go forward (2 * (1 + 1)) end
  ]
]
\end{lstlisting}
\end{minipage}
\subsubsection{Syntactic Style Variations}

We generate the same underlying AST in different surface forms to test robustness to syntax.
The \texttt{Block} style uses explicit keywords like \texttt{do}/\texttt{end} and \texttt{repeat} to delimit scope.

\noindent 
\begin{minipage}{\linewidth}
\begin{lstlisting}[style=example]
# Block Style
run ((0 + 1) * 4) times {
  if ((has cube plus_and has key_1) plus_and (has box_4 or has key_1)) after {
    do go forward 2 end
  }
}
\end{lstlisting}

The \texttt{C-style} mimics C-like syntax with curly braces \texttt{\{\}} and semicolons.

\begin{lstlisting}[style=example]
// C-Style
run(((0 + 1) * 4)){
  if(((holding(cube) and holding(key_1)) and (holding(box_4) alt holding(key_1)))){
    go(forward,2);
  }
}
\end{lstlisting}
\end{minipage}

The \texttt{S-expr} style uses Lisp-like prefix notation with fully parenthesized expressions.

\noindent 
\begin{minipage}{\linewidth}
\begin{lstlisting}[style=example]
; S-Expression Style
(repeat (* (+ 0 1) 4)
  (if (and (and (has cube) (has key_1)) (or (has box_4) (has key_1))) next
    (go forward 2)
  )
)
\end{lstlisting}
\end{minipage}

\subsubsection{Lexical Familiarity Variations}

We vary the lexicon to disentangle syntactic reasoning from semantic priors.
\texttt{Natural} lexicon uses familiar English keywords.

\noindent 
\begin{minipage}{\linewidth}
\begin{lstlisting}[style=example]
# Natural Lexicon
if ((has item_3 and has box_4) and not (has ball_1)) after [
  if ((has ball_4 and has ball) and (has box_0 alt has key)) after [
    repeat (0 + (1 * 0)) x [
      repeat 4 x [ ... ]
    ]
  ]
]
\end{lstlisting}
\end{minipage}

\texttt{Alien} lexicon replaces keywords with randomized nonsense tokens, stripping semantic cues.

\noindent 
\begin{minipage}{\linewidth}
\begin{lstlisting}[style=example]
# Alien Lexicon
v_pipo ((v_gsqe item_3 v_vvyj v_gsqe box_4) v_vvyj v_tkik (v_gsqe ball_1)) v_oklp [
  v_pipo ((v_gsqe ball_4 v_vvyj v_gsqe ball) v_vvyj (v_gsqe box_0 v_lwmz v_gsqe key)) v_oklp [
    v_zmew (0 v_vyht (1 v_lbke 0)) v_buty [
      v_zmew 4 v_buty [ ... ]
    ]
  ]
]
\end{lstlisting}
\end{minipage}

\subsection{Typical Failure Modes and Case Studies}
\label{sec:appendix-failure-modes}

Following the three-layered evaluation hierarchy, we provide abstracted case studies to localize specific reasoning failures. For clarity of presentation, the examples below utilize the \texttt{Natural} lexicon style rather than the \texttt{Alien} mode to ensure the underlying logic remains interpretable. Furthermore, to accommodate the structural complexity of programs at extreme depths, we present only the critical segments of the code where errors occur. Non-essential instructions are omitted to focus exclusively on the breakdown of syntactic, behavioral, or semantic reasoning.

\subsubsection{Syntactic Fragility}

\paragraph{Case Study 1: Lexical Mapping Failure}
This failure identifies a symbolic binding gap where the model tracks the AST structure but fails to replace EBNF non-terminal with their terminal strings.

\noindent 
\begin{minipage}{\linewidth}
\begin{lstlisting}[style=errbox, escapechar=|]
# [Model Output]
|\hlerr{DO MOVE}| forward 5 |\hlerr{END}| 
|\hlerr{IF}| ( |\hlerr{NOT}| ( |\hlerr{HOLDING}| item_2 ) ) ...
|\hlerr{THEN}| [ |\hlerr{DO GRAB}| cube_3 |\hlerr{END}| ] 

# [Ground Truth]
do move forward 5 stop 
if (not (has item_2)) ...
next [ do take cube_3 stop ] 
\end{lstlisting}
\end{minipage}

\paragraph{Case Study 2: Hierarchical Scoping Failure}
This failure demonstrates a breakdown in recursive state-tracking, where the model applies a flat encapsulation heuristic to hierarchical structures.

\noindent 
\begin{minipage}{\linewidth}
\begin{lstlisting}[style=errbox, escapechar=|]
# [Model Output]
run 3 iters {
  |\hlerr{exec}| if (no has box_0) next { 
    exec go backward 5 fin
  } |\hlerr{fin}|
}

# [Ground Truth]
run 3 iters {
  if (no (has box_0)) next { 
    exec go backward 5 fin
  }
}
\end{lstlisting}
\end{minipage}

\paragraph{Case Study 3: Lexical Operator Confusion}
This failure occurs when the model fails to translate arithmetic descriptors into the formal mathematical operators required by the grammar, despite correctly identifying the calculation logic.
\noindent 
\begin{minipage}{\linewidth}
\begin{lstlisting}[style=errbox, escapechar=|]
# [Model Output]
exec go forward (3 |\hlerr{times}| 2) fin
exec go backward ((2 |\hlerr{times}| 4) |\hlerr{plus}| (1 |\hlerr{plus}| 3)) fin

# [Ground Truth]
exec go forward (3 * 2) fin
exec go backward ((2 * 4) + (1 + 3)) fin
\end{lstlisting}
\end{minipage}

\paragraph{Case Study 4: Premature Scope Termination}
This failure occurs when the model loses track of deep nesting levels, causing it to close an outer control block prematurely or omit trailing delimiters as the sequence length increases.

\noindent 
\begin{minipage}{\linewidth}
\begin{lstlisting}[style=errbox, escapechar=@]
# [Model Output]
if (has item_1) {
    if (no (has box_2)) {
        ...
    }
@\hlerr{\}} @
@\hlerr{// Missing second brace '\}'}@

# [Ground Truth]
if (has item_1) {
    if (no (has box_2)) {
        ...
    }
}
\end{lstlisting}
\end{minipage}

\subsubsection{Behavioral Misalignment}
\paragraph{Case Study 1: Logical Predicate Inversion}
This failure occurs when a model produces syntactically valid code that adheres to all EBNF rules but fails to faithfully simulate the conditional logic of the instructions. The model correctly identifies the predicates but substitutes the wrong logical connective, leading to an incorrect execution path.

\noindent 
\begin{minipage}{\linewidth}
\begin{lstlisting}[style=errbox, escapechar=@]
# [Model Output]
when (not (has ball)) @\hlerr{plus\_and}@ ((has key_3 plus_and has key_4)) next [
    ... 
]

# [Ground Truth]
when (not (has ball) @\hlerr{alt}@ (has key_3 plus_and has key_4)) next [
    ... 
]
\end{lstlisting}
\end{minipage}

\paragraph{Case Study 2: Arithmetic Simulation Collapse}
This failure occurs when the model generates syntactically correct expressions but fails to accurately simulate the underlying mathematical or procedural state, leading to a divergence in the final environment state.

\noindent 
\begin{minipage}{\linewidth}
\begin{lstlisting}[style=errbox, escapechar=@]
# [Model Output]
exec go backward @\hlerr{(4 * 5 + 1)}@ end 

# [Ground Truth]
exec go backward (4 * 5 * 1) end
\end{lstlisting}
\end{minipage}

\paragraph{Case Study 3: Planning Horizon Collapse}
This failure identifies a collapse where the model successfully navigates a high-complexity sub-task but fails to resume the global execution plan, leaving the robot stranded before reaching the goal.

\noindent 
\begin{minipage}{\linewidth}
\begin{lstlisting}[style=errbox, title={Abstracted Behavioral Truncation}, escapechar=@]
# [Model Output (Syntactically Valid)]
if (has box plus_and has key) then {
    if no (has item_2) then { ... }
} 
@\hlerr{// ERROR: Program terminates here}@

# [Ground Truth / Expected Behavior]
if (has box plus_and has key) then { ... }
act move forward 5 fin  
\end{lstlisting}
\end{minipage}

\subsubsection{Semantic Drift}

\paragraph{Case Study 1: Arithmetic AST Flattening}
This failure identifies a "computation-over-alignment" bias where the model prioritizes behavioral success over structural adherence. While the resulting program achieves the correct environment state, it fails the \textbf{Semantic Correctness} check because the model simplifies complex expressions into literal results, discarding the original AST structure.

\noindent 
\begin{minipage}{\linewidth}
\begin{lstlisting}[style=errbox, escapechar=@]
# [Model Output]
# The model evaluates the expression instead of preserving its structure
run @\hlerr{19}@ iters {
    exec turn left end
}

# [Ground Truth]
run ((4 * 4) + 3) iters {
    exec turn left end
}
\end{lstlisting}
\end{minipage}

\paragraph{Case Study 2: Boolean AST Flattening}
This failure occurs when the model correctly identifies all logical predicates but collapses the hierarchical nesting of a boolean expression into a single-level chain. 

\noindent 
\begin{minipage}{\linewidth}
\begin{lstlisting}[style=errbox, title={Abstracted Boolean Structure Merging}, escapechar=@]
# [Model Output]
if @\hlerr{(not}@ @\hlerr{holding}@ @\hlerr{item\textunderscore 1)}@ @\hlerr{plus\textunderscore and}@ @\hlerr{(holding}@ @\hlerr{cube\textunderscore 4)}@ @\hlerr{plus\textunderscore and}@ @\hlerr{(holding}@ @\hlerr{box\textunderscore 1)}@ then {
    ...
}

# [Ground Truth]
if (not (holding item_1) plus_and (holding cube_4 plus_and holding box_1)) then {
    ...
}
\end{lstlisting}
\end{minipage}

\paragraph{Case Study 3: Structural Redundancy via Autoregressive Echoing} This failure stems from the nature of autoregressive decoding, where previously generated complex blocks act as strong distractors in the context. The model's attention mechanism may loop by re-predicting the same high-probability sequence instead of advancing the execution plan. While behaviorally idempotent, this redundancy creates a structural mismatch with the ground-truth AST.

\noindent 
\begin{minipage}{\linewidth}
\begin{lstlisting}[style=errbox, title={Abstracted Structural Redundancy}, escapechar=@]
# [Model Output]
if no ( ... ) then { run 4 times { ... } } 
@\hlerr{if no ( ... ) then \{ run 4 times \{ ... \} \}}@ 
do turn left fin

# [Ground Truth]
if no ( ... ) then { run 4 times { ... } } 
do turn left fin
\end{lstlisting}
\end{minipage}
\end{document}